 \documentclass[12pt,letterpaper]{article}		

\usepackage{leipzig}
\usepackage{tipa} 
\usepackage{graphicx}
\usepackage{linguex}

\usepackage{enumitem}
\setlist{noitemsep, topsep=6pt} 

\usepackage{times}
\usepackage[sort]{natbib}	
\setcitestyle{comma,aysep={},yysep={,},notesep={; }}
\usepackage{fullpage} 
\usepackage[compact]{titlesec}
	\titleformat{\section}[runin]{\normalfont\bfseries}{\thesection.}{.5em}{}[.]
	\titleformat{\subsection}[runin]{\normalfont\scshape}{\thesubsection.}{.5em}{}[.]
	\titleformat{\subsubsection}[runin]{\normalfont\scshape}{\thesubsubsection.}{.5em}{}[.]
\usepackage[usenames,dvipsnames]{color}	
\definecolor{splinkcolor}{rgb}{.0,.2,.4}
\usepackage[colorlinks,allcolors={black},urlcolor={splinkcolor}]{hyperref} 		

\usepackage[vskip=6pt,leftmargin=0.3in,rightmargin=0.3in,indentfirst=false]{quoting}

\makeatletter
\def\@maketitle{%
  \newpage
  \begin{center}%
  \let \footnote \thanks
    {\normalfont\bfseries \@title \par}%
    {\vskip .5em
      \begin{tabular}[t]{c}%
        \@author
      \end{tabular}\par}%
  \end{center}%
  \par}
\makeatother
\renewenvironment{abstract}{%
\noindent\begin{minipage}{1\textwidth}
\setlength{\leftskip}{0.4in}
\setlength{\rightskip}{0.4in}
\textbf{Abstract.}}
{\end{minipage}}
\newenvironment{keywords}{%
\vspace{.5em}
\noindent\begin{minipage}{1\textwidth}
\setlength{\leftskip}{0.4in}
\setlength{\rightskip}{0.4in}
\textbf{Keywords.}}
{\end{minipage}}

\usepackage[font={small}]{caption}
\usepackage{hyperref,soul,breakurl,expex,microtype}

\title{What can LLMs tell us about the mechanisms behind polarity illusions in humans? Experiments across model scales and training steps}
\author{Dario Paape\footnote{Author: Dario Paape, Department of Linguistics, University of Potsdam (\href{mailto:paape@uni-potsdam.de}{paape@uni-potsdam.de}).}}

\begin{document}

\widowpenalty10000
\clubpenalty10000
\sloppy
 
\maketitle
\thispagestyle{empty}
\begin{abstract}
I use the Pythia scaling suite \citep{biderman2023pythia} to investigate if and how two well-known polarity illusions, the NPI illusion and the depth charge illusion, arise in LLMs. The NPI illusion becomes weaker and ultimately disappears as model size increases, while the depth charge illusion becomes stronger in larger models. The results have implications for human sentence processing: it may not be necessary to assume ``rational inference'' mechanisms that convert ill-formed sentences into well-formed ones to explain polarity illusions, given that LLMs cannot plausibly engage in this kind of reasoning, especially at the implicit level of next-token prediction. On the other hand, shallow, ``good enough'' processing and/or partial grammaticalization of prescriptively ungrammatical structures may both occur in LLMs. I propose a synthesis of different theoretical accounts that is rooted in the basic tenets of construction grammar.
\end{abstract}


\vspace{6pt} 

\section{Introduction} 

The sentences in (\ref{x1}a,b) can create so-called linguistic illusions, where a grammatical error is easy to miss in certain structural configurations:

\pex\label{x1}
\a The shareholders that no executives misled have \underline{ever} filed a suit.
\a No detail is too small to be \underline{missed}.
\xe

Both of these sentences involve negation, and both can create confusion regarding the polarity of their component parts. Sentence (\ref{x1}a) may generate a negative polarity item (NPI) illusion: the NPI \emph{ever} cannot be licensed by the negative quantifier \textit{no} in the embedded relative clause, yet the sentence sometimes appears grammatical to speakers (e.g., \citealp{parker2016negative,drenhaus2005processing}). Sentence (\ref{x1}b) is an example of the ``depth charge'' illusion. The degree phrase \textit{too small to be missed} is inherently incongruous; if anything, a detail may be too small to be \textit{noticed}. The presence of the negative quantifier \textit{no} at the beginning of the sentence should not ``repair'' this incongruity, yet speakers are often under the impression that it does, and that \textit{missed} is negated when it is actually not (e.g., \citealp{sanford2002depth}). The depth charge illusion is one of the strongest known grammaticality illusions, and often proves highly resistant to correction \citep{wason1979verbal}, to the extent that it has even been argued to be an example of grammaticalization (e.g., \citealp{cook2010no}). By contrast, the NPI illusion is usually argued to be transient, and is easier to spot when the two constructions are presented to the same speakers \citep{paape2024linguistic}.

Different theories have been proposed about the emergence of the NPI illusion and the depth charge illusion in human speakers. I will focus on three broad classes of theories that I will call the \textit{shallow processing account}, the \textit{rational inference account}, and the \textit{grammaticalization account}. The shallow processing account in its most general form states that readers in some way fail to correctly apply their mental grammar when linguistic illusions occur. According to \citet[p.~382]{sanford2002depth}, shallow processing means that ``each word in a sentence does not necessarily contribute its full meaning, and these meanings are not always combined into higher-level phrase meanings through a fully determinate analysis''. When processing is shallow, the sentence processor may resort to simple heuristics instead of rigorously applying grammatical rules (e.g., \citealp{christianson2016language}). In the case of (\ref{x1}a), this might mean that \textit{ever} fails to check whether \textit{no} is in a c-commanding position, and in the case of (\ref{x1}b), it might mean that \textit{no} is simply assumed to negate \textit{missed}, which has the side effect of yielding a sensible interpretation.

The rational inference account takes quite a different approach. It assumes that grammar \textit{is} rigorously applied, but that the speaker's representation of the literal sentence does not always match the actual string after it was transmitted through the ``noisy channel'' of real-world communication \citep{levy2008noisy,gibson2013rational}. Under this model, the processor effectively ``glosses over'' inconsistencies that are likely to be typos or speech errors, and the speaker may believe that \textit{no} is or was supposed to be in a position that c-commands \textit{ever} in (\ref{x1}a) \citep{muller2022could}, or that the string in (\ref{x1}b) was actually \textit{No detail is so small as to be ignored} \citep{zhang2023noisy}. Such mental ``edits'' are assumed to be more likely the more semantically and syntactically plausible the non-veridical analysis is, and the more ``noisy'' the current processing environment is \citep{ryskin}.

A grammaticalization-based account has, to my knowledge, only been proposed for the depth charge illusion. Given the aforementioned stability of the illusion even in the face of explicit correction attempts, some authors have argued that the template \textit{No X is too to Z} is a stored grammatical construction with an idiomatic, non-compositional meaning in which the final verb is negated \citep{cook2010no,fortuin2014deconstructing}. By contrast, for NPI illusions, apart from the observation that they are usually amenable to explicit correction, treating them as grammaticalized would have several undesirable consequences, such as failing to explain their less-than-full acceptability, and also casting doubt on otherwise well-supported theories of NPI licensing \citep{muller2022could}. Nevertheless, it is possible in principle that some speakers actually believe that (\ref{x1}a) is grammatical.

Despite their very different assumptions, it has proven difficult to find experimental evidence that supports one of the three accounts over the others. For instance, all three accounts assume that the plausibility of the illusory interpretation affects the strength of the illusion. All three accounts also predict that illusion sentences should be easier to process than their clearly ungrammatical counterparts --- that is, sentences that completely lack negation --- assuming that ``repairs'' via rational inference happen non-consciously with no or little cognitive effort, and that shallow processing is faster than deep processing \citep{paape2024linguistic}. Experimental data including reading times and acceptability judgments, as well as reported effects of the strength about prior assumptions on illusion strength \citep{zhang2023noisy,paape2020quadruplex}, thus do not offer a way of deciding between the three theories. In fact, there is a possibility that all three accounts are correct to some degree.

Large language models (LLMs) offer a potential solution to the empirical puzzle. For the present purpose, I will intentionally limit the discussion to LLMs that are trained only on text with a next-word prediction objective, and that have \textit{not} been augmented through mechanisms such as information retrieval, dialogue optimization via human feedback, or ``reasoning'' mechanisms via repeated self-prompting (e.g., \citealp{guu2020retrieval, ouyang2022instructions,shojaee2025illusion}). These ``bare'' LLMs learn statistical regularities from the linguistic input, which enables them to infer a probability distribution over upcoming words (or rather, tokens, which may be part-words).

Whether LLMs learn actual syntactic rules from the statistical regularities they are exposed to during training remains debated. To some extent, LLMs may leverage shallow heuristics instead of detailed syntactic representations to predict upcoming tokens. For example, the heuristic ``multiple singular nouns are likely to be followed by a plural verb'' often correctly predicts verbal agreement in coordination structures (e.g., \citealp{mueller-linzen-2023-plant,marvin2018targeted}). Making use of heuristics to meet the prediction objective instead of applying complex syntactic rules would suggest similarities between LLM processing and the shallow processing sometimes seen in humans \citep{cong2025language}. In particular, for the NPI and depth charge sentences in (\ref{x1}), an LLM could fall back on ``there is a negation somewhere in the sentence'', similar to what has been suggested for human speakers \citep{paape2020quadruplex,muller2022could}. It is also possible that during training, an LLM would first learn superficial heuristics and then at some point transition to abstract syntactic rules, this becoming immune to the illusions over time. Likewise, larger LLMs may show less reliance on heuristics due to their increased capacity.

While human shallow processing has a plausible analogue in LLMs, this is highly questionable in the case of rational inference about plausible transmission errors \citep{paape2023transformer}. Given their nature as statistically-driven word predictors without access to any experience beyond language, it is highly unlikely --- and arguably downright impossible --- that (especially ``bare'', unaugmented) LLMs can ``reason'' about likely errors and ``correct'' them, which would, inter alia, require some semblance of a human-like theory of mind and of metacognition, which even the most recent systems demonstrably do not possess \citep{marchetti2025artificial,huff2025judgments}. However, for the current purpose, this shortcoming is an advantage: if rational inference about ``noisy'' communication architecturally cannot arise in LLMs but LLMs still show the same linguistic illusions as humans for (\ref{x1}a,b), this would suggest that invoking a rational inference process is not necessary to explain them.

The predictions of the grammaticalization account diverge for the NPI illusion and the depth charge illusion in LLMs. Crucially, these predictions relate to the plausible learning trajectories and scaling behaviors across different model sizes that should arise. If \textit{No X is too Y to Z} is indeed a grammaticalized construction, it is a rare one \citep{fortuin2014deconstructing}. Thus, we might hypothesize that an LLM would first learn to complete (\ref{x1}b) compositionally, that is, with a ``positive'' verb like \textit{noticed} (or \textit{seen}, or \textit{cherished}, ...) as opposed to a ``negative'' verb like \textit{missed} (or \textit{left out}, ...), because it has not picked up on the idiosyncratic meaning of the construction yet. However, as the model keeps seeing more input data during training, it should potentially pick up on the pattern and start producing \textit{missed} or similar ``negative'' completions. Larger models should also potentially show more \textit{missed}-like completions because they can encode more information, and may thus be more sensitive to the unique properties of the construction. For the NPI illusion, the models may start out by using simple heuristics, and later during training pick up on the correct licensing mechanism. Larger models may also be more likely to learn the correct syntactic rule.

\section{LLM sentence completion experiments}

The Pythia scaling suite \citep{biderman2023pythia} offers a unique testing ground for the present research question. The Pythia suite contains 16 transformer-based LLMs of different sizes (70 million to 12 billion parameters) that are all trained on the same data, namely the Pile \citep{gao2020pile}.\footnote{For the present experiments, the models trained on the deduplicated version of the Pile were used.} Furthermore, for each model, the suite provides checkpoints that are evenly spaced across the training process, so that the learning curve of each model can be relatively closely followed.

Previous studies on NPI and depth charge illusions in LLMs yielded mixed results. \citet{shin2023investigating} and \citet{zhang2023can} found the NPI illusion across different transformer-based LLMs, in the sense that the NPI \textit{ever} was predicted more strongly in sentences with an embedded \textit{no} than in sentences without it. \citet{paape2023transformer} and \citet{zhang2023can} also found the depth charge illusion to be present in several transformer models, but as the models struggled with some of the control conditions that also involved negation and degree phrases, the results are not as straightforward to interpret. Furthermore, unlike for the NPI illusion, the models showed a much weaker depth charge illusion than humans, in the sense that they often did not have a strong prediction for a negative verb (\textit{No detail is too small to be \underline{missed}}) and predicted positive verbs instead, which is compatible with a compositional, non-illusory reading of the sentence. The interpretability of these previous findings is also limited by the fact that the log probability of a predefined word (\textit{ever}, \textit{missed}) was used as the dependent measure.\footnote{\citet{zhang2023can} also used whole-sentence perplexity, but this measure suffers from the same caveats.} Differences in the critical word's log probability between conditions are, of course, informative, but may be confounded with differences in entropy: the models' probability distribution may simply become more spread out across different words, which does not automatically entail that, across all possible continuations, negative-polarity completions become more likely than positive-polarity ones.

Given these caveats, the current experiments use a wider array of control conditions for the depth charge illusion, and focus on the abstract semantic properties of the LLMs' predictions rather than on specific words. To achieve this, I use beam search to find the top 50 three-token continuations for each prompt. For each prompt, the top 50 most likely next tokens are sampled, then for each of these, the 50 most likely \textit{next} tokens are sampled, pruning is applied, and the process is applied once more for the third token. The resulting top 50 three-token continuations are then categorized according to the following scheme:

\begin{itemize}
\item For NPI sentences: Continuations are categorized as negative if they contain any NPI and do \textit{not} themselves contain a licensing negation. Possible NPIs include \textit{ever}, members of the \textit{any} family (e.g., \textit{anyone}, \textit{anything}), as well as others that were incrementally added to the list after checking completions (e.g., \textit{in weeks}, \textit{at all}, \textit{yet}).\\[-2ex]
\item For depth charge sentences: Continuations are categorized as negative if they contain ``negative'' verbs (e.g., \textit{missed}, \textit{left out}, \textit{ignored}) or occasionally adjectives (\textit{meaningless}). An exact definition of what makes a verb (or other word) ``negative'' in the context of the depth charge illusion is difficult to give, but it is generally agreed that the best examples express an absence of action and/or a negative attitude towards the subject of the sentence \citep{paape2020quadruplex,kizach2016verbal}.
\end{itemize}

The categorization step was carried out by the author with some assistance from Claude.ai (Anthropic PBC)\footnote{Claude.ai was \textit{not} used to write or edit any part of this paper, nor were any other LLM-based tools.} and two additional human judges. To compute the final dependent measure, the log probabilities of all continuations within the positive and negative categories were exponentiated and summed for each prompt, and the resulting summed probabilities were transformed back to the log scale.\footnote{Zero probabilities were smoothed by adding 0.001 before transformation.} The difference between the summed log probabilities of the negative category versus the positive category ($\Delta$logprob) then indicates ``how negative'' the continuations were overall.

\paragraph{Materials} 72 NPI preambles and 64 depth charge preambles were used in the experiment. 24 NPI sentences were taken from \citet{muller2022could} while the rest were newly created. For the depth charge sentences, 32 were taken from \citet{zhang2023noisy} while the rest were newly created.

NPI preambles appeared in three conditions (\textit{no}, \textit{embedded-no}, and \textit{the}), and always ended with the word \textit{have}:

\begin{center}
\begin{minipage}{0.7\textwidth}
\begin{itemize}
\item[\fbox{\textit{no}}] ~~~\textit{No shareholders that the executives misled have~\ldots}\\[-2ex]
\item[\fbox{\textit{emb-no}}] ~~~\textit{The shareholders that no executives misled have~\ldots}\\[-2ex]
\item[\fbox{\textit{the}}] ~~~\textit{The shareholders that the executives misled have~\ldots}\\[-2ex]
\end{itemize}
\end{minipage}
\end{center}

Depth charge preambles appeared in eight conditions as shown below, and always ended with the word \textit{be}. The \textit{no} condition is the classic depth charge configuration, whereas the \textit{this} condition is the classic control condition without a negative quantifier, which only very rarely generates an illusion in humans. The other conditions are control conditions designed to probe the models' general handling of negation and degree phrases. The continuations shown in square brackets are example continuations that would be expected under a compositional reading of the sentence.

\begin{center}
\begin{minipage}{0.7\textwidth}
\begin{itemize}
\item[\fbox{\textit{no}}] ~~~\textit{No detail is too small to be~\ldots}\hfill[noticed]\\[-2ex]
\item[\fbox{\textit{this}}] ~~~\textit{This detail is too small to be~\ldots}\hfill[noticed]\\[-2ex]
\item[\fbox{\textit{this not}}] ~~~\textit{This detail is not too small to be~\ldots}\hfill[noticed]\\[-2ex]
\item[\fbox{\textit{too}}] ~~~\textit{This detail is too small and should be~\ldots}\hfill[ignored]\\[-2ex]
\item[\fbox{\textit{too not}}] ~~~\textit{This detail is too small and should not be~\ldots}\hfill[noticed]\\[-2ex]
\item[\fbox{\textit{simply}}] ~~~\textit{Very small details can simply be~\ldots}\hfill[ignored]\\[-2ex]
\item[\fbox{\textit{very}}] ~~~\textit{Very small details should usually be~\ldots}\hfill[ignored]\\[-2ex]
\item[\fbox{\textit{very not}}] ~~~\textit{Very small details should usually not be~\ldots}\hfill[noticeable]
\end{itemize}
\end{minipage}
\end{center}

As the examples show, there is a natural grouping of conditions: compositionally, the \textit{no}, \textit{this}, \textit{this not}, \textit{too not} and \textit{very not} conditions should be positive while the \textit{too}, \textit{simply} and \textit{very} conditions should be negative. Intuitively, the presence and strength of the depth charge illusion can be determined by whether the \textit{no} condition groups more with the positive or with the negative prompts.

\paragraph{Models} For the scaling component of the experiment, the final checkpoints of all Pythia models from the 160-million-parameter model up to the 12-billion-parameter model were used. For the training component, the 1.4-billion-parameter model was chosen as it represents the midpoint of the scaling sample. A convenience sample of checkpoints was selected that was intended to cover the entire training process, with somewhat more emphasis on earlier and later compared to intermediate training steps (256, 512, 1.000, 2.000, 10.000, 50.000, 100.000, 143.000 [final]). All models were accessed via Hugging Face libraries \citep{wolf2019huggingface}.

\begin{figure}[htb]\centering
\begin{center}
\begin{minipage}{0.8\textwidth}
\includegraphics[scale=.5]{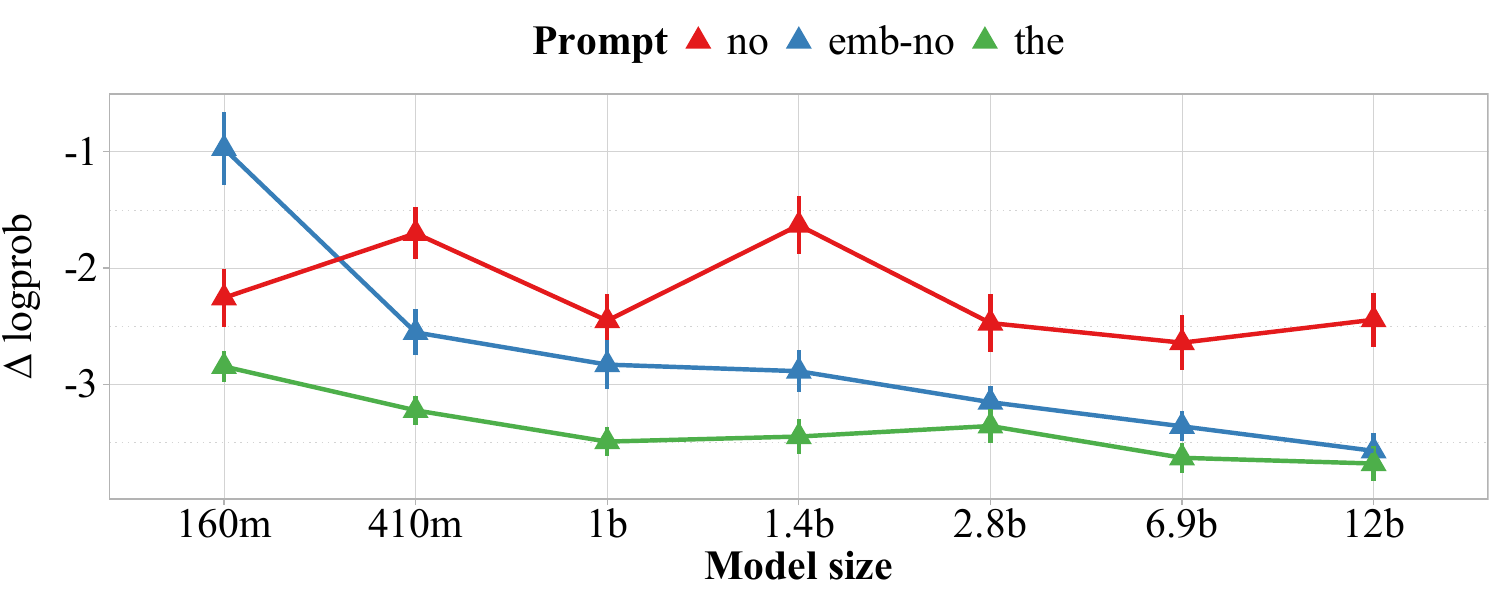}
\caption{Difference in summed log probabilities between negative and positive continuations for NPI sentences across prompts and model sizes. More positive y-values mean more negative continuations. Error bars show 95\% confidence intervals.\label{npis}}
\end{minipage}
\end{center}
\end{figure}

\paragraph{Results for NPI sentences} Figure \ref{npis} shows results for the NPI scaling component of the experiment. The results are relatively straightforward: The smallest Pythia model shows a pattern in which an embedded \textit{no} generates more NPIs than a matrix-clause \textit{no}, presumably due to a proximity heuristic. Starting with the next-smallest model, this pattern flips, but the \textit{emb-no} condition still generates more NPIs than the \textit{the} condition, indicating the presence of an NPI illusion. As the models grow larger, this illusion gradually disappears. The largest model shows no NPI illusion at all.

\begin{figure}[htb]\centering
\begin{center}
\begin{minipage}{0.8\textwidth}
\includegraphics[scale=.5]{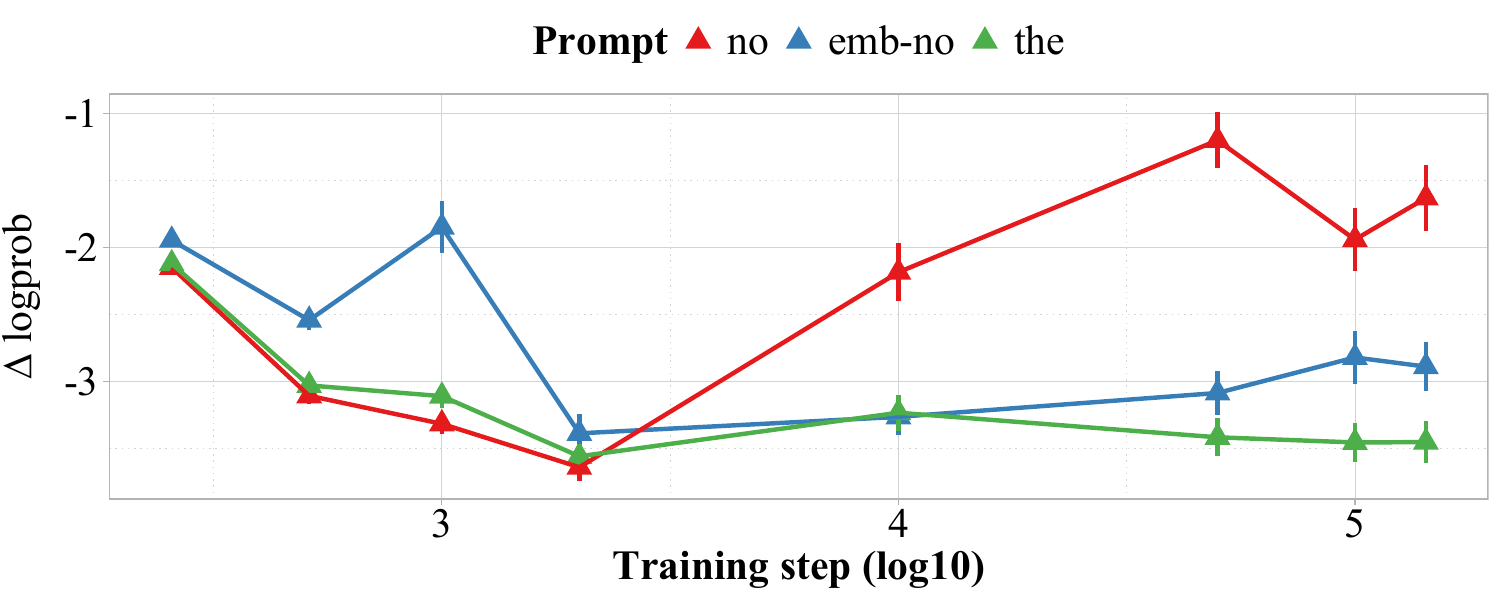}
\caption{Difference in summed log probabilities between negative and positive continuations for NPI sentences in the 1.4b model across prompts and training steps. More positive y-values mean more negative continuations. Error bars show 95\% confidence intervals.\label{npit}}
\end{minipage}
\end{center}
\end{figure}

Figure \ref{npit} shows results from the 1.4b model for the NPI training component of the experiment. This model shows an early peak of many NPIs being generated in the \textit{emb-no} condition around 1.000 training steps. At training step 10.000, the model shows more NPIs in the \textit{no} condition, and no NPI illusion in the \textit{emb-no} condition. However, the classic NPI illusion pattern (\textit{no} $\textgreater$ \textit{emb-no} $\textgreater$ \textit{the}) starts to appear in later training steps.

\begin{figure}[htb]\centering
\begin{center}
\begin{minipage}{0.8\textwidth}
\includegraphics[scale=.5]{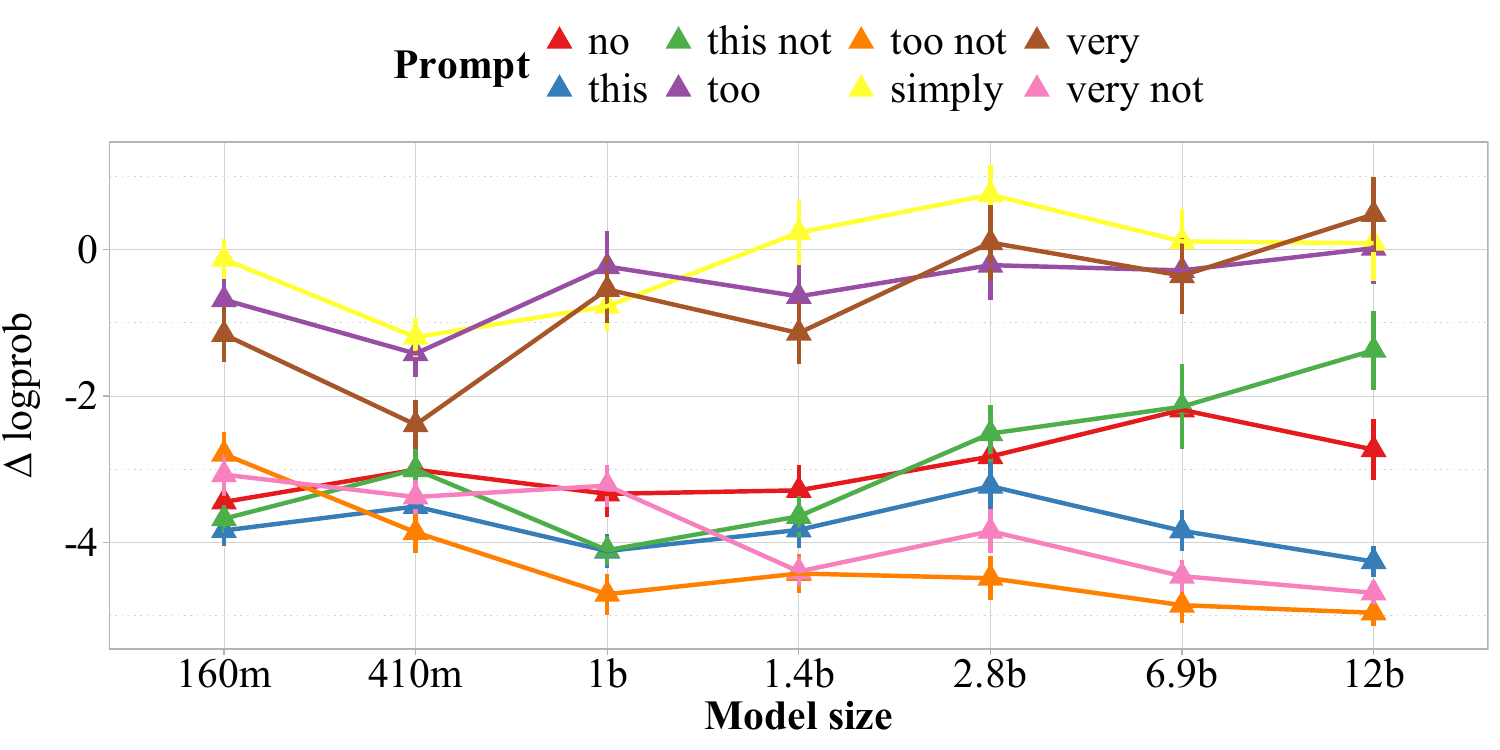}
\caption{Difference in summed log probabilities between negative and positive continuations for depth charge sentences across prompts and model sizes. More positive y-values mean more negative continuations. Error bars show 95\% confidence intervals.\label{dcs}}
\end{minipage}
\end{center}
\end{figure}

\paragraph{Results for depth charge sentences} Figure \ref{dcs} shows results for the depth charge scaling component of the experiment. Overall, the positive constructions (see above) become more positive and the negative constructions become more negative for larger models. There are, however, two exceptions to this pattern: the classic depth charge configuration (\textit{no}) groups with positive constructions for smaller models but moves towards the negative constructions for larger models. The result for the 12b model may indicate a reversal of this trend for even larger models, but this interpretation is speculative. 

The second exception is the \textit{this not} construction, which clearly tends more negative for larger models. This result is somewhat surprising. The \textit{this not} construction is compositionally positive but the word \textit{not} is close to the end of the preamble, which may suggest a proximity-based heuristic that results in negative continuations. However, why the larger models should make \textit{more} use of such a hypothesized heuristic is not clear.

\begin{figure}[htb]\centering
\begin{center}
\begin{minipage}{0.8\textwidth}
\includegraphics[scale=.5]{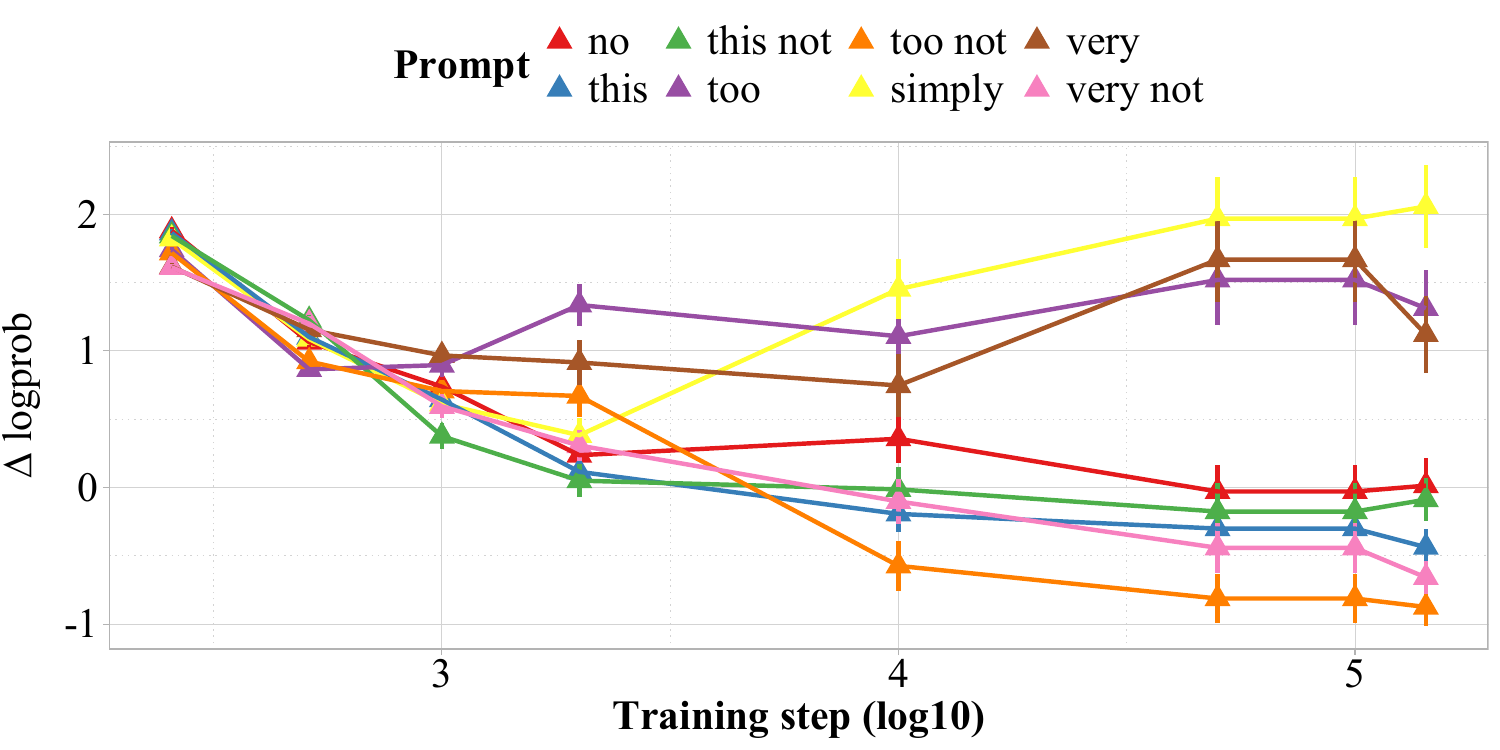}
\caption{Difference in summed log probabilities between negative and positive continuations for depth charge sentences in the 1.4b model across prompts and training steps. More positive y-values mean more negative continuations. Error bars show 95\% confidence intervals.\label{dct}}
\end{minipage}
\end{center}
\end{figure}

Figure \ref{dct} shows results from the 1.4b model for the depth charge training component of the experiment. For this model, the positive and negative constructions start to diverge around training step 10.000, and the \textit{no} and \textit{this not} constructions pattern with the positive constructions.

\section{Discussion} The aim of the present experiments on the NPI and depth charge polarity illusions was to use data from LLMs to learn something about the plausible origin of these illusions in humans. I have argued above that rational inference about plausible speech errors and other input ``corruptions'' is unlikely to be within the scope of what ``bare'' LLMs can achieve, and yet LLMs are susceptible to both types of polarity illusion. If polarity illusions can occur without rational inference, by Occam's razor, one may conclude that it is also not necessary to assume such a mechanism in humans, at least not as the primary driver behind these types of effects. The current results are somewhat nuanced, however, as scale and training of the LLM do seem to matter: the largest Pythia model was immune to the NPI illusion but was susceptible to the depth charge illusion. In this model, depth charge sentences patterned somewhere between the positive and negative constructions, unlike in humans, for whom depth charge sentences overwhelmingly pattern negative. Furthermore, some of the smaller models are largely immune to the depth charge illusion, which constitutes an even more striking contrast with humans. Overall, a possible tentative conclusion is that while it is not necessary to assume a rational inference mechanism to explain polarity illusions in humans, such a mechanism could potentially explain some of the mismatches between human and LLM behavior. I will return to this point below.

The shallow processing account assumes that capacity or motivation limits (e.g., \citep{christianson2022if}) drive human speakers to sometimes use heuristics to construct impoverished but subjectively ``good enough'' representations of linguistic input. Insufficient motivation is not a plausible factor for LLMs, and they also lack the memory bottleneck that requires humans to process input as quickly as possible before new input arrives \citep{christiansen2016now}. However, LLMs \textit{do} need to be efficient encoders, and may take shortcuts during training that lead to statistically plausible but nevertheless incorrect generalizations. One example of this in the current data could be the observation that the smallest Pythia model, as well as the 1.4b model early during training, showed an NPI illusion that is completely unattested in humans, where embedded \textit{no} generates more NPIs than matrix \textit{no}, presumably because the embedded \textit{no} is closer to the critical continuation point. 

The scaling behavior of the NPI illusion is in line with the assumption that the larger the models become, the more they are able to encode complex syntactic constraints, and the less susceptible they are to the illusion. By contrast, the scaling behavior of the depth charge illusion appears to go against this pattern, given that the larger models tend to be more rather than less susceptible to it. Overall, the results are compatible with similar patterns of shallow processing in LLMs and humans if the correct NPI licensing rule but not the correct interpretation rule for depth charge sentences is learnable by the LLMs in the Pythia suite. As pointed out by \citet{paape2023transformer}, the absence of the depth charge illusion in smaller models could due to their use of a simple heuristic: by ignoring the beginning the sentence and only focusing on the degree phrase (\textit{too small to notice}), the compositionally correct verb can be relatively easily predicted.

While it has received far less attention in the literature, the grammaticalization account also turns out to be a strong contender in terms of providing an explanation for the scaling and training patterns of the tested LLMs, and possibly also for the illusion patterns observed in humans. Recall that grammaticalization has been proposed for the depth charge illusion \citep{fortuin2014deconstructing,cook2010no} but not for the NPI illusion. Assuming that this dichotomy captures a real contrast, it is to be expected that the NPI illusion disappears at sufficiently large model sizes while the depth charge illusion gets stronger. Grammaticalization of the depth charge construction is also in line with the ``illusion'' being famously resistant to correction, as well as extremely difficult to detect in error-spotting tasks \citep{wason1979verbal,paape2024linguistic}.

To my mind, the overall pattern of results suggests a synthesis of all three accounts, with some additional inspiration from construction grammar and ``usage-based'' approaches to language (e.g., \citealp{goldberg2010construction}). Constructions are conceived of as form-meaning pairings that are learned from language experience. Constructions that are similar in form and meaning will overlap with each other in the learner's conceptual space and form clusters, and similar constructions can mutually influence each other's acquisition and processing. It has recently been argued that LLMs are a perfect example of purely usage-based learning, and that their acquired representations also behave in the way predicted by construction grammar \citep{goldberg2024usage,ambridge2024large}. Construction grammar shares some theoretical assumptions with the shallow or ``good enough'' processing framework, most importantly the assumptions that incorrect constructions can sometimes be accessed during processing and that ``blended'' constructions can result in analyses and interpretations that prescriptively should not be possible \citep{barlow2000usage,malyutina2016lingers,frazier2015without,goldberg2022good}. It has also been argued that language acquisition in humans may involve a progression from shallow to deeper processing \citep{ferrara2025children}, which may be true for LLMs as well. 

Regarding a connection between construction grammar and rational inference mechanisms, one candidate concept is the idea of ``near neighbor'' sentences that the rational inference framework uses to formalize the mental edits that speakers presumably make to change ill-formed utterances into a form that is more amenable to grammatical analysis (e.g., \citealp{levy2008noisy,keshev2021noisy}). These edits are conceived of as edits to the actual linguistic input, but perhaps they are simply interference from neighboring constructions that influence the processing of a given utterance. In combination with heuristic processing strategies that survive well past initial acquisition, and additional influences of background knowledge and perhaps metalinguistic skills in humans, such interference effects may create phenomena like the depth charge illusion, which is mostly indistinguishable from full grammaticalization.

As a final remark, I want to stress that despite the parallels between LLMs and humans in terms of susceptibility to polarity illusions that the current study has focused on, correlation between human and LLM-derived data is not a compelling argument in favor of some ``deep'' human-LLM similarity in terms of cognition more generally \citep{guest2023logical}. Moreover, the current study has focused on a single family of transformer-based language models, the Pythia suite. Even across different transformers, results within a single grammatical construction may vary widely, and may differ in varying ways from human performance \citep{vondermalsburg2026diverging}. Nevertheless, by taking into account which cognitive abilities ``base'' LLMs \textit{cannot} plausibly have, such as reasoning in any proper sense, we can still learn something about human cognition by pruning away or reconceptualizing theoretical constructs that are not needed to explain the observed behavior (e.g., \citealp{michaelov2023peanuts}). Thus, even though I heartily agree that ``LLMs don't know anything'' \citep{goddu2024llms}, I would still argue that if a stochastic parrot \citep{parrots} can do a given trick, no complex machinery may be necessary for a human to perform the same trick.

\setlength{\bibhang}{0.3in}			
\titleformat{\section}{\normalfont\bfseries}{\thesection}{.5em}{}

\bibliographystyle{elsarticle-harv}		
\newcommand{\doi}[1]{\href{#1}{#1}}	
\vspace{6pt}

\bibliography{pythia}

\end{document}